\documentclass[lettersize,journal]{IEEEtran}
\usepackage{bbding}
\usepackage{amsmath,amsfonts}
\usepackage{algorithmic}
\usepackage{array}
\usepackage[caption=false,font=normalsize,labelfont=sf,textfont=sf]{subfig}
\usepackage{textcomp}
\usepackage{stfloats}
\usepackage{url}
\usepackage{multirow}
\usepackage{verbatim}
\usepackage{graphicx}
\usepackage{booktabs}
\usepackage[breaklinks]{hyperref}
\usepackage{balance}

\def\BibTeX{{\rm B\kern-.05em{\sc i\kern-.025em b}\kern-.08em
    T\kern-.1667em\lower.7ex\hbox{E}\kern-.125emX}}

\begin{document}
\title{Unsupervised Low Light Image Enhancement\\
	via SNR-Aware Swin Transformer}

\author{\IEEEauthorblockN{Zhijian Luo\thanks{* Zhijian Luo is the corresponding author, Email: \href{mailto:luozhijian@jyu.edu.cn}{luozhijian@jyu.edu.cn}},
Jiahui Tang,
Yueen Hou,
Zihan Huang and
Yanzeng Gao\\}
\IEEEauthorblockA{School of Computer, Jiaying University, \\
Meizhou, R. P. China, 514015\\}
}


\maketitle

\begin{abstract}
Image captured under low-light conditions presents unpleasing artifacts, which debilitate the performance of feature extraction for many upstream visual tasks.
Low-light image enhancement aims at improving brightness and contrast, and further reducing noise that corrupts the visual quality.
Recently, many image restoration methods based on Swin Transformer have been proposed and achieve impressive performance.
However, on one hand, trivially employing Swin Transformer for low-light image enhancement would expose some artifacts, including over-exposure, brightness imbalance and noise corruption, etc.
On the other hand, it is impractical to capture image pairs of low-light images and corresponding ground-truth, i.e. well-exposed image in same visual scene.
In this paper, we propose a dual-branch network based on Swin Transformer, guided by a signal-to-noise ratio prior map which provides the spatial-varying information for low-light image enhancement.
Moreover, we leverage unsupervised learning to construct the optimization objective based on Retinex model, to guide the training of proposed network.
Experimental results demonstrate that the proposed model is competitive with the baseline models.
\end{abstract}


\section{Introduction}
\IEEEPARstart{H}{igh-energy} visibility images contain abundant information about the target scene, which is crucial for most visually-based tasks including object detection \cite{sarin2019face,liu2016ssd}, image classification \cite{al2022comparing}, and image denoising \cite{anaya2018renoir}, among other classic upstream visual tasks.
In the process of image acquisition, there are often many uncontrollable physical factors, which degenerate the quality of captured image and consequently debilitate the performance of feature extraction for many upstream visual tasks.
Among these factors, imaging in low-light environment is very common in daily life, therefore low-light image enhancement(LLIE) has become more and more urgent for visualization and many computer vision task.\cite{chen2022survey}.


\par The Retinex theory\cite{land1971lightness}, which was proposed by Land and McCann in 1971, provided a visual and intuitive explanation for the imaging process of  low-light image.
The primary hypothesis is that the low-light image can be decomposed to reflectance and illumination.
Many Retinex-based methods\cite{jobson1997multiscale, jobson1997properties} following \cite{land1971lightness} tried to estimate the illumination at the first stage, and restore the reflectance as the final result according to the estimated illumination.
Although the details could be to some extent restored from low-light image, many artifacts including ringing and over-exposed occur in enhanced image.
To address these issues, several model-based methods\cite{guo2016lime, li2018structure} have been proposed in which different priors were manually designed as regularization terms to model the characteristics of illumination and reflectance.
However, model-based optimization often involves the process of iteration, which is relatively time-consuming, hence is limited to apply in practice.

\par Over the past of decade, deep learning has been shown superiority in many low-level computer vision tasks, and has been successfully applied in LLIE\cite{chen2018learning, guo2020zero, ren2020lr3m, jiang2021enlightengan, wei2018deep, zhang2021self, xu2022snr}.
Among these methods, Retinex-based methods\cite{wei2018deep, zhang2021self, zhang2021unsupervised} employ deep networks, whose backbone are based on the convolution neural  networks(CNNs), to jointly estimate the components of reflectance and illumination.
However, globally employing same convolution kernel on entire image can not extract information in regions under different light conditions. In other words, it is independent on the content of image.
Furthermore, long-range dependency can not be modeled due to the local modeling of convolution kernel, hence resulting in the loss of global information.

\begin{figure}
	\centering
	\includegraphics[width=1\linewidth]{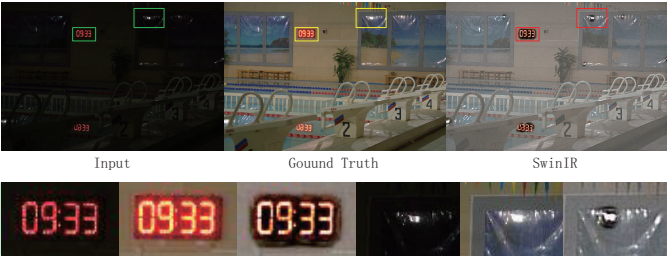}
	\caption{Using SwinIR as backbone for Low-light image enhancement. From the comparison of image details, it can be seen that there are many unpleasing artifacts in directly applying SwinIR for LLIE, including over-exposure, noise corruption and brightness imbalance. For example, estimations via SwinIR in regions of numerical clock and the top left corner of window reveal artifacts of brightness imbalance and noise corruption. Over-exposure occurs in restoration of SwinIR at the top left corner of window.} \label{fig:fig1}
\end{figure}

In order to model the long-range dependency, recently a backbone  based on Swin Transformer named SwinIR was proposed\cite{liang2021swinir}, and achieved outstanding performance on many vision tasks, including super-resolution\cite{luo2022bsrt}, image segmentation\cite{lin2022ds}, object detection\cite{liu2022swin}, and image denoising\cite{zhang2022practical}, etc.
In this paper, we also consider Swin Transformer as our main backbone. However, as shown in Fig.\ref{fig:fig1}, trivially using SwinIR to enhance low-light image leads to some unpleasing artifacts, including over-exposure, noise corruption and brightness imbalance.
These annoying artifacts reveal that SwinIR could not be able to distinguish the levels of light conditions in low-light image.
An signal-to-noise ratios(SNR)-guided transformer network was proposed in \cite{xu2022snr} based on the observation that regions of higher/lower SNR typically have higher/lower visibility and less/more noise.
However, the training of network is based upon the framework of supervised learning, where image pairs of low-light and normal-light in the same visual scene are required, and the data collection is rather time-consuming and expensive.
Moreover, there may be multiple low-light and high-light images in the same scene, so it can be difficult to determine the best image from reference image.
Despite receiving expert correction, the process of selecting the optimal reference image can remain a formidable challenge.
To eliminate the dependencies on pair of training data, a series of methods\cite{zhang2021unsupervised, jin2022unsupervised, zhang2021self, zhang2020self} based on unsupervised learning have been proposed.


\par In this paper, we proposed a dual branch Network (ASW-Net) with signal-to-noise ratio aware Swin Transformer for LLIE.
The proposed ASW-Net is composed of two branches, shallow feature extraction and deep feature extraction module.
Within the deep feature extraction module, long dependency of image is modeled via Swin Transformer guided by SNR map.
On the other hand, for regions with relatively high SNR, they have more valid information features, and local information in this region is sufficient for image enhancement.
Hence, the shallow Feature Extraction Module is based on residual convolution neural network\cite{simonyan2014very}.
Overall, the main contributions of our work are three-folds:
\begin{enumerate}
	\item We propose a dual branch network with SNR-aware Swin Transformer, named ASW-Net, for low-light image enhancement. To address some annoying artifacts while trivially using SwinIR to enhance low-light image, including over-exposure, noise corruption and brightness imbalance, we design an SNR-aware transformer module for achieving spatial-varying low-light image enhancement.
	\item To eliminate the dependence of data collection and to generalize the network ASW-Net to be extended to enhance images collected from different scenes, different lighting conditions and different devices, we build an unsupervised framework based on Retinex model, possessing more feasible and efficient training of network.
	\item Extensive experiments on two benchmark datasets demonstrate that the proposed ASW-Net is competitive with the state-of-the-art baselines in terms of PSNR and SSIM evaluation metrics.
\end{enumerate}

\section{Related Work}
The classic Retinex theory\cite{jobson1997multiscale} provides a clear and intuitive description of physical process of LLIE.
This theory suggests that the perception of an object's color and brightness of the human visual system depends on the reflection characteristics of the object's surface, while the reflectance of an object under different lighting conditions can be perceived by the HVS.
The low-light image can be decomposed into reflectance component and illumination component,
\begin{equation}
	S=R\circ I \label{Eq:Retinex_noiseless}
\end{equation}
where $S \in \mathbb{R}^{H\times W \times 3}$ denotes the observed low-light image, $R \in \mathbb{R}^{H\times W \times 3}$ and $I \in \mathbb{R}^{H\times W}$ are the component of reflectance and illumination respectively, and $\circ$ is the element-wise production.
The enhancement of LLIE tries to estimate both components jointly.
However, in practical, the imaging of low-light image is often complicated due to the presence of noise and other artifacts.

Li \textit{et al}.\cite{li2018structure} proposed a robust Retinex model, which establish optimization objectives by defining different prior constraints.
However, since that prior constraints are insufficient, it often produces result with over-smoothing and insufficient brightness.
The method(LR3M) proposed by Ren \textit{et al}. \cite{ren2020lr3m} tries to apply low-rank prior in Retinex decomposition process to suppress noise, which has also been successfully applied to low-light video enhancement and achieved excellent results.
Furthermore, Kong \textit{et al}.\cite{kong2021low} proposed a Poisson noise-aware Retinex model, which preserves image structure information and reduces noise in the meantime.
On the whole, the core ingredient of most model-based methods is to design priors that model the characteristics of illumination and reflectance.
However, prior constraints often rely on assumptions in the real-world environment, and the representational capacity of priors is limited.
Moreover, the optimization of model-based methods often involves numerical iterations, which are relatively time-consuming and not conducive to practical applications.

Over the past of decade, deep learning has been shown superiority in many low-level computer vision tasks, and the development of  LLIE has been significantly promoted by deep learning technology.
Chen \textit{et al}.\cite{wei2018deep} proposed a RetinexNet based on deep Retinex decomposition, which enhances brightness using an enhancement network for lightness and performs denoising operations on reflectivity.
Although this method is effective at improving image brightness but can result in artifacts and distortions due to weak constraints on intermediate variables.
Zhang \textit{et al}.\cite{zhang2019kindling} proposed a low-light image enhancer named KinD, by introducing a series of illumination constrains, and it successfully solved the phenomena of over-exposure.
Obviously, current LLIE techniques have achieved a methodological leap from traditional model design to data-driven deep learning at the methodological level.


Considering the insufficient generalization performance of the existing paired data training mechanism and the inaccuracy of the paired data itself, a series of methods have been proposed to mitigate the dependence on paired data. Jiang \textit{et al}.\cite{jiang2021enlightengan} developed a Generative Adversarial Network with a self-attention mechanism, and trained it in an unpaired fashion. Guo \textit{et al}.\cite{guo2020zero} constructed a pixel-level curve estimation convolution neural network called Zero-DCE through iterative derivation, and designed a series of zero-reference training loss functions to solve the low-light image enhancement problem. These methods still require high correlation between the training samples with the test images in terms of image content and noise statistics.

\section{Method}

\begin{figure*}
	\centering
	\includegraphics[width=0.9\linewidth]{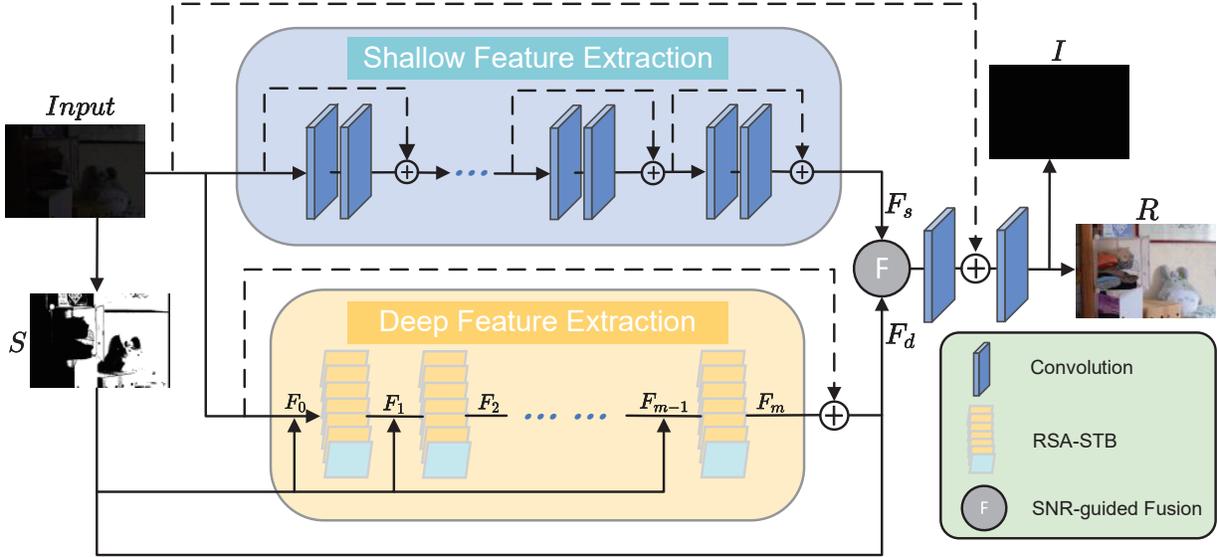}
	\caption{Overview architecture of our proposed ASW-Net. ASW-Net is a dual branch network where one branch is shallow feature extraction module, and the other branch is deep feature extraction module. The final estimation outputted by the last convolution layer contains illumination and reflectance which is considered to be the desired normal-light image.}
	\label{Fig_architecture}
\end{figure*}

To suppress the unpleasing artifacts of over-exposure, noise corruption and brightness imbalance, we proposed an unsupervised framework for low-light image enhancement
of a dual branch network with SNR-aware Swin Transformer, named ASW-Net.
The overall architecture of ASW-Net is presented in Fig.\ref{Fig_architecture}.

As shown in Fig.\ref{Fig_architecture}, ASW-Net is a dual branch network where one branch is shallow feature extraction module, and the other branch is deep feature extraction module.
The first branch is based on convolution neural network, while the other is based on residual SNR-aware Swin Transformer block (RSA-STB).
Based on observation that regions of higher/lower SNR typically have higher/lower visibility and less/more noise, shallow feature extraction module employs convolution structure to capture local information which plays core role in regions of high SNR, while deep feature extraction module adopts Swin Transformer guided by SNR map to capture non-local messages which are adequate in regions of very low SNR.
Furthermore, enhancement of each pixel should be aware of the contribution of local and non-local information.
To achieve this, following \cite{xu2022snr}, we employ an SNR-aware fusion module to guide the fusion of shallow feature map $F_s$ and deep feature maps $F_d$ as shown in Fig.\ref{Fig_architecture}.
Following the fusion module, a sequential structure containing the residual connection and a convolution layer is utilized to generate the final estimation consisting of illumination and reflectance maps.
We choose the reflectance map as the desired solution, i.e., normal-light image.

In the same scene, there may be multiple low-light and high-light image pairs so that it is hard to determine the best image from reference images even with expert correction, making it difficult to obtain pairwise training data.
Furthermore, to guarantee the generalization of the network, i.e., could extend to enhance images collected from different scenes, different brightness conditions and different devices, we employ the unsupervised learning to train our ASW-Net.

\subsection{Shallow Feature Extraction Module}
Traditional methods of low-light image enhancement mostly use convolution structures as backbone of the network, and these methods mainly focus on capturing the local information of the image.
For image regions of high visibility which have relatively high SNR, the regions of local information plays a major role.
Hence, compared with non-local information, local information is more accurate to restore in these regions.
We constructs a shallow feature extraction module based on convolution residual blocks as show in Fig. 2.
We let $F_s \in \mathbb{R}^{H\times W \times C}$ denotes the output of shallow feature extraction module.

\subsection{Deep Feature Extraction Module}
Compared with CNNs model, Transformer model have competitive performance, such as image classification\cite{dosovitskiy2020image}.
However, directly using transformer to vision tasks has a problem: the transformer has the drawback of modeling long sequences because input sequence is the entire pixel of image. To address this issue, the Swin Transformer model\cite{liu2022swin} was proposed.
It solves the problem efficiency by using a shifting window approach, resulting in a reduction of parameters required still achieving state-of-the-art performance.

However, as show in Fig. 1, trivially using Swin Transformer to enhance low-light image leads to some unpleasing artifacts, including over-exposure, noise corruption and brightness imbalance. These annoying artifacts reveal that Swin Transformer could not be able to distinguish the levels of light conditions in low-light image.
Image regions with extremely low SNR contain a large amount of noise\cite{xu2022snr, chandler2007vsnr}, and it is insufficient to enhance only by local information.
Hence, we propose an SNR-aware Swin Transformer layer for enhancing image regions of extremely low SNR, which is based on observation that regions of higher/lower SNR typically have higher/lower visibility and less/more noise.

\begin{figure}[!ht]
	\centering
	\includegraphics[width=1\linewidth]{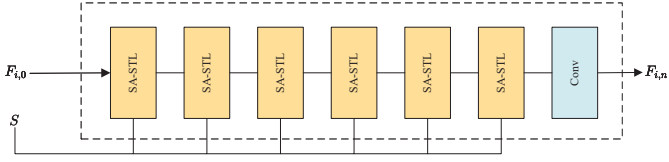}
	\caption{Residual SNR-Aware Swin Transformer Blocks (RSA-STB).}
\end{figure}

\subsubsection{Residual SNR-aware Swin Transformer Block}
As show in Fig. 3, the residual SNR-aware Swin Transformer block(RSA-STB) is consisting of multiple SNR-Aware Swin Transformer layers(SA-STL) and a $3\times 3$ convolutional layer.
Specifically, given the input feature $\mathcal{F}_{i-1} \in \mathbb{R}^{H\times W\times C}$ of the $i$-th RSA-STB, the process of the $i$-th RSA-STB is formulated as:
\begin{equation}
	\left\{
	\begin{split}
		& F_{i,0}=\mathcal{F}_{i-1} \\
		& F_{i,j}=\textit{SA-STL}(F_{i,j-1}, S) \\
		& F_{i,n+1}=\textit{Conv}(F_{i,n}) + F_{i,0}\\
		& \mathcal{F}_i=F_{i,n+1},~~ \text{and} \hspace{0.2cm} i=1,2,3 \ldots m, j=1,2,3 \ldots n,
		\label{RSA-STB}
	\end{split}
	\right.
\end{equation}
where \textit{SA-STL}$(\cdot,\cdot)$ is a single SNR-Aware Swin Transformer layer and takes the output of last SA-STL and the SNR map as inputs, $j$ denotes the layer index of SA-STL, \textit{Conv}$(\cdot)$ is the convolution layer in $i$-th RSA-STB, $\mathcal{F}_i \in \mathbb{R}^{H\times W\times C}$ is the output of the $i$-th RSA-STB and $S$ is the SNR map of input image, which is defined in section \ref{sec_SNR}.
After $m$ RSA-STB, the output is $\mathcal{F}_m$ which is further combined with $\mathcal{F}_0$ to represent the deep feature $F_d$, i.e.,
\begin{equation}
	F_d = \mathcal{F}_0 + \mathcal{F}_m.
\end{equation}

\subsubsection{The SNR Map} \label{sec_SNR}
We firstly convert input image $ I \in \mathbb{R}^{H\times W\times 3} $ to a grayscale image $I'\in \mathbb{R}^{H\times W} $.
It is challenging to directly calculate the SNR value for each pixel from input image $I$.
In some non-learning based denoising methods \cite{buades2005image,chandler2007vsnr,chen2019seeing}, the noise component can be model as distance between noise image and noise-free image, i.e,
\begin{equation}
	E= abs(I'- denoise(I')),\label{Estimation}
\end{equation}
where $abs$ is the absolute value, $denoise$ is denoising operation.
Then we compute $\hat{S}$ the ratio of grayscale image $I'$ and the noise component $E$ as follow
\begin{equation}
	\hat{S}=\frac{I'}{E},
\end{equation}
and the approximation of the SNR map $S \in \mathbb{R}^{H \times W}$ is obtained by normalizing the values of $\hat{S}$ to range $[0,1]$, i.e.,
\begin{equation}\label{SNR map}
	S_{i,j} = \left\{
	\begin{split}
		&0, ~~ \text{if}~~ \hat{S}_{i,j}<0 \\
		&\hat{S}_{i,j}, ~~\text{if}~~ 0\leq \hat{S}_{i,j} \leq 1 \\
		&1, ~~\text{otherwise}
	\end{split}
	,
	\right.
\end{equation}
where $i\in \{1,\cdots,H\}, j\in \{1,\cdots, W\}$.

\subsubsection{SNR-aware Swin Transformer Layer}
\begin{figure*}[!t]
	\centering
	\subfloat[The flowchart of SA-STL.]
	{\includegraphics[width=0.6\textwidth]{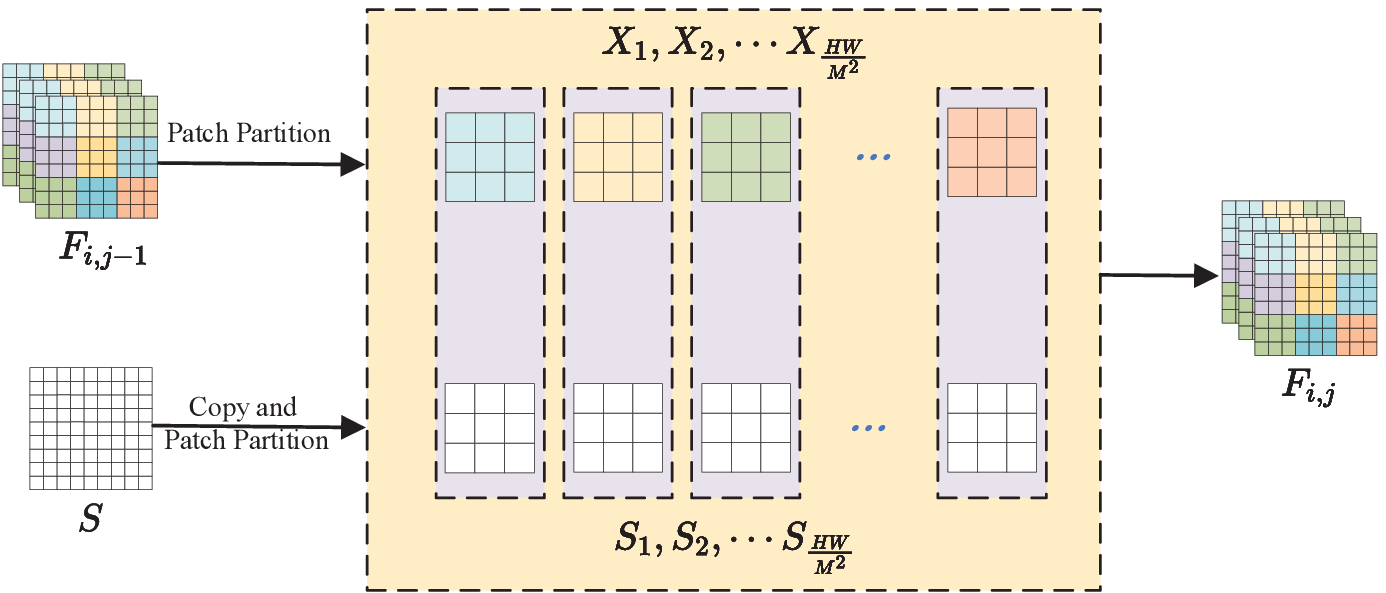 } } \hspace{3em}
	\subfloat[SNR-Aware self-attention for each local window]
	{\includegraphics[width=0.12\textwidth]{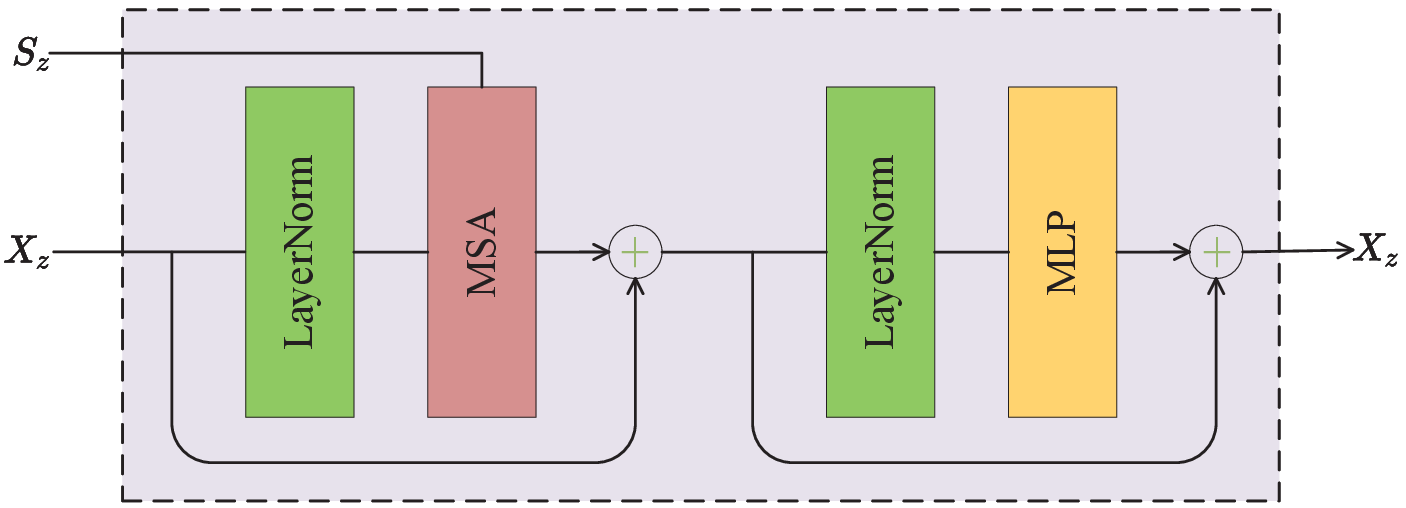}}
	\caption{SNR-Aware Swin Transformer layer(SA-STL).}
	\label{Fig_SA_STL}
\end{figure*}

In the last of this section, we state the details of SA-STL which is shown in Fig.\ref{Fig_SA_STL}.
In order to extract long dependency through Swin Transformers for enhancing images, the signal and noise levels should be taken into consideration \cite{xu2022snr}.
Based on the observation that regions of higher/lower SNR typically have higher/lower visibility and less/more noise, the enhancement of low-light image via long-range attention should be guided by the levels of noise, i.e., the SNR map.
Hence, during the attention computation of Swin Transformer, we leverage the SNR map as a mask to shake off the risk of message propagation from regions of low SNR \cite{xu2022snr}.
The mask value of $S$ is set to be
\begin{equation}
	S_{i j}=\left\{\begin{array}{l}
		0, S_{i j} \leq t \\
		1, S_{i j} \geq t
	\end{array}, i \in \{1, \cdots, H\}, j=\{1, \cdots, W\}\right. .\label{Normalize}
\end{equation}
where $t$ is a threshold and is set to be $0.5$ in our experiments.

Given an input feature $F_{i,j-1} \in \mathbb{R}^{H \times W \times C}$ of the $j$-th SA-STL in the $i$-th RSA-STB, following SwinIR \cite{liang2021swinir} we reshape the input feature $F_{i,j-1}$ to a size of $\frac{HW}{M^2} \times M^2 \times C$ by dividing the feature map into non-overlapping local windows with size of $M \times M$, where $\frac{HW}{M^2}$ indicates the number of local windows.
In order to guide the attention of Swin Transformer by SNR map, we first stack $C$ copy of $S$ to form tensor $S' \in \mathbb{R}^{H \times W \times C}$, and also reshape $S'$ to a size of $\frac{HW}{M^2} \times M^2 \times C$ by partitioning it into local windows to match the size of $F_{i,j} \in \mathbb{R}^{\frac{HW}{M^2} \times M^2 \times C}$.
Then as shown in Fig. \ref{Fig_SA_STL}(a), the SNR-Aware self-attention is computed based on the standard multi-head self-attention \cite{vaswani2017attention} in parallel for each window as follow.

For each local window, the inputs of SNR-aware self-attention are 2D feature $X_z \in \mathbb{R}^{M^2 \times C}$ and corresponding local SNR map $S_z \in \mathbb{R}^{M^2 \times C}$, where $z=1,\cdots ,\frac{HW}{M^2}$, and $X_z$ and $S_z$ are the $z$-th slice of tensor $F_{i,j-1}$ and $S'$ respectively.
Then the computation of query matrix $Q_z$, key matrix $K_z$, and value matrix $V_z$ are as follow:
\begin{equation}
	Q_z=X_z W_q,K_z=X_z W_k,V_z=X_z W_v,
\end{equation}
where $W_q$, $W_k$, and $W_v$ represent the projection matrices which share across different windows.

Suppose that we have $Q_z,K_z,V_z \in \mathbb{R}^{M^2 \times d}$, where $d$ is the dimension of projection space, and the attention matrix of each local window is expressed as
\begin{equation}
	\begin{split}
		&\text{Attention}(Q_z,K_z,V_z,S_z)\\
		&~~=\text{softmax} \left(\frac{Q_z K_z^T}{\sqrt{d} } +B+(1-S_z)(1-S_z)^T \delta \right)V_z,
	\end{split}
	\label{attention}
\end{equation}
where $B$ is a learnable matrix of relative position encoding, and $\delta$ is a small negative constant $-1e9$.
In practice, we implement the attention function Eq.\eqref{attention} based on the multi-head self-attention, and combine the results by concatenating them for each head of self-attention.
As show in Fig. \ref{Fig_SA_STL}(b), the whole process of SNR-aware self-attention for each window $z$ is formulated as:
\begin{equation}
	\begin{split}
		& X_z = MSA(LN(X_z), S_z) + X_z, \\
		& X_z = MLP(LN(X_z)) + X_z,    \label{multhead attention}
	\end{split}
\end{equation}
where $MLP(\cdot)$ is a multi-layer perceptron which is consisted of two fully-connected layers and $LN(\cdot)$ is the LayerNorm layer.

After that, given $X_z$ the outputs of SNR-aware self-attention for each local window, we combine these outputs by merging them to form the output feature $F_{i,j}$.
To enable cross-window connections of SNR-aware self-attention for each window, regular and shifted window partition are employed alternately \cite{liu2022swin}.

\subsection{Fusion Module}
As shown in Fig.\ref{Fig_architecture}, through shallow and deep feature extraction module, we obtain the shallow feature $F_s \in \mathbb{R}^{H\times W \times C}$ and the deep feature $F_d \in \mathbb{R}^{H\times W \times C}$.
To encourage the enhancement of each pixel to be aware of the contribution of $F_s$ and $F_d$, we employ the fusion module to combine $F_d$ and $F_s$ guided by SNR map $S$.
The fused feature map $F_{f} \in \mathbb{R}^{H\times W\times C}$ is determined dynamically based on the SNR map.
Formally, the fused feature map $F_{f}$ is obtained by interpolating $F_s$ and $F_d$, i.e.,
\begin{equation}
	F_{f}=F_s \times S+ F_d \times (1-S). \label{Feature-Fusion}
\end{equation}

After that, we adopt a convolution layer to transfer the fused feature map $F_f$ with size of ${H\times W\times C} $ into residual with size of ${H\times W\times 3}$.
Following the residual connection, a convolution layer is employed to estimate the final feature map $F_{out} \in \mathbb{R}^{H\times W\times 4}$ whose first $3$ channels form the component of reflectance $R \in \mathbb{R}^{H\times W\times 3}$ and last channel is that of illumination $I \in \mathbb{R}^{H\times W}$.

\subsection{Loss Function for Unsupervised Learning}
In the same scene, there may be multiple low-light and high-light image pairs so that it is hard to determine the best image from reference images even with expert correction, making it difficult to obtain pairwise training data.
Hence, the collection of data sample is rather time-consuming and expensive.
Furthermore, it is also doubtable whether the pre-trained model could be extended to enhance images collected from different scenes, different lighting conditions and different devices, i.e. how to ensure the generalization of this model.
To address these issues, unsupervised learning is often employed to eliminate the highly dependence on paired data and boost the generalization of model.

According to the Retinex theory, an image can be decomposed into reflectance map and illumination map, which is expressed in Eq.\eqref{Eq:Retinex_noiseless}.
However, due to the presence of noise in the low-light image, the degradation of low-light image is mathematically modeled as
\begin{equation}
	S= R\circ I + N
	\label{Eq:Retinex_noise}
\end{equation}
where $N$ represents some unknown noise during degradation of low-light image.
Reflectance component delineates the intrinsic property of object in normal light, and is often regarded to be consistent under any lightness condition in some work\cite{zhang2020self,wei2018deep}.

Given the degraded low-light image $S$, the enhancement is equivalent to maximizing the posterior $p(R,I|S)$.
According to Bayesian theorem, the posterior probability could be expressed as
\begin{equation}
	p(R,I| S) \propto p(S| R,I)\cdot p(R) \cdot p(I), \label{Bayes}
\end{equation}
where $p(S| R,I)$ is likelihood probability, and $p(R)$ and $p(I)$ are the priors of reflectance and illumination respectively.
Hence from the point of statistical view, the problem of low-light image enhancement is equivalent to maximizing the posterior $p(R,I| S)$ with respect to $R$ and $I$, i.e,
\begin{equation}\label{Eq:Bayse_Objective}
	\max_{R,I} p(S| R,I)\cdot p(R) \cdot p(I).
\end{equation}
In this paper, we suppose that $N$ is additive Gaussian noise, i.e., $ N \sim \mathcal{N}(0,\sigma^2 I)$, which indicates
\begin{equation}
	p(S|R,I) = \frac{1}{ \left( 2 \pi \sigma^2 \right)^{\frac{H \times W \times 3}{2}}  } \exp \left\{ -\frac{\left\|S- R\circ I \right\|_F^2}{2\sigma^2} \right\},
\end{equation}
where $\|\cdot\|_F$ denotes the Frobenius norm and $P(S|R,I) = P(N)$ holds.

By taking the negative logarithm of Eq.\eqref{Eq:Bayse_Objective}, the problem of low-light image enhancement can be transformed into a minimization problem including data term and regularization terms as:
\begin{equation} \label{Eq:Optimization_Objective}
	\min_{R,I} \frac{1}{2} \left \|S- R\circ I \right\|_F^2 + \lambda_1 \ell_R + \lambda_2 \ell_I,
\end{equation}
where $\ell_R$ and $\ell_I = $ are regularization terms of reflectance and illumination respectively, and $\lambda_1$ and $\lambda_2$ are weight parameters that control the trade-off between data term and regularization terms.
Obviously, from Eq.\eqref{Eq:Bayse_Objective} and \eqref{Eq:Optimization_Objective}, $-\log p(R) = \lambda_1 \ell_R$ and $-\log p(I) = \lambda_2 \ell_I$ hold.

\subsubsection{Prior of Reflectance}
Based upon the fact that image are locally smooth and the intensity of pixel gradually varies in most regions, total variation(TV)-based regularization \cite{rudin1992nonlinear} is commonly used for suppressing noise in the task of LLIE \cite{park2017low, fu2015probabilistic}.
Regarding the prior of reflectance, we also adopt TV regularization, i.e., the $L_1$ norm of the gradient of reflectance $\| \nabla R \|_1$, where $\nabla$ denotes gradient operator.
However, TV regularization would lead the textures to be over-smooth and stair-casing phenomenon takes place in flat region which are approximated by piecewise constant surface.

Since the maximum value of image channel has great impact on visual effect of image, we consider that the maximum value of the reflectance should comply with that of low-light image.
In order to lighten the dark image while preserving the maximum amount of information from the low-light image, following \cite{zhang2020self,zhang2021self}, we uses histogram equalization to improve the entropy of the processed image.
Based on these, the regularization of reflectance component $R$ is formulated as:
\begin{equation}
	\ell_R=\left \| \max_{c\in r,g,b}  R^{c}- F \left( \max_{c\in r,g,b}  S^{c} \right) \right \|_1 +\lambda \left \| \nabla R \right \| _1, \label{Reflection_Loss}
\end{equation}
where $F(\cdot)$ is the histogram equalization operator, $\lambda$ is a weight parameter, and $ \max_{c \in r,g,b} R^{c}$ represents the maximum channel value of the RGB image $R^{c}$.

\subsubsection{Prior of Illumination}
Regarding the prior of illumination, the illumination map should be local consistent and structure aware, i.e., it should be smooth in textural details while can preserve the structure boundary of objective.
To this end, we utilize the structure-aware and smooth regularization on this map proposed in \cite{wei2018deep}:
\begin{equation}
	\ell_I=\left \| \nabla I\circ \exp\left(-\lambda_3 \nabla R \right) \right \|_1, \label{Lightness_Loss}
\end{equation}
where $\lambda_4$ is a weight parameter balancing the trade-off between smoothness and structure-awareness.

\subsubsection{Unsupervised leaning}
Based on Eq. \eqref{Eq:Optimization_Objective}, \eqref{Reflection_Loss} and \eqref{Lightness_Loss}, we can derive the overall objective function to train the ASW-Net network as
\begin{equation} \label{Eq:Objective}
	\begin{split}
		L  =&
		\frac{1}{2}\left \| S-R\circ  I  \right \|_F^2
		+ \lambda_1 \left \| \max_{c\in R,G,B} R^{c}- F\left( \max_{c\in R,G,B}  S^{c} \right) \right \|_1 \\
		& + \lambda_2 \left \| \nabla I\circ \exp\left(-\lambda_3 \nabla R \right) \right \|_1
		+\lambda_4\left \| \nabla R \right \|_1.
	\end{split}
\end{equation}

Optimization-based methods typically involve multiple iterations of numerical process to solve the above minimization problem, which is considerable time-consuming and challenging, due to some complex constraints introduced in Eq. \eqref{Eq:Objective}.
Therefore, it is tough for enhancing of low-light image in real time via solving above minimization problem in numerical way.
In order to tackle this problem, learning-based approach is commonly utilize via designing proper network to approximate the mapping of enhancement from low-light image to well-exposed one.
However, the training of network is based on the framework of supervised learning, where image pairs of low-light and normal-light in the same visual scene are required, and data collection is rather time-consuming and expensive.
For unsupervised learning, the network for LLIE is trained on explicit loss function which contains only low-light image as shown in Eq. \eqref{Eq:Objective}.

\begin{figure*}[!ht]
	\centering
	\includegraphics[width=1\linewidth]{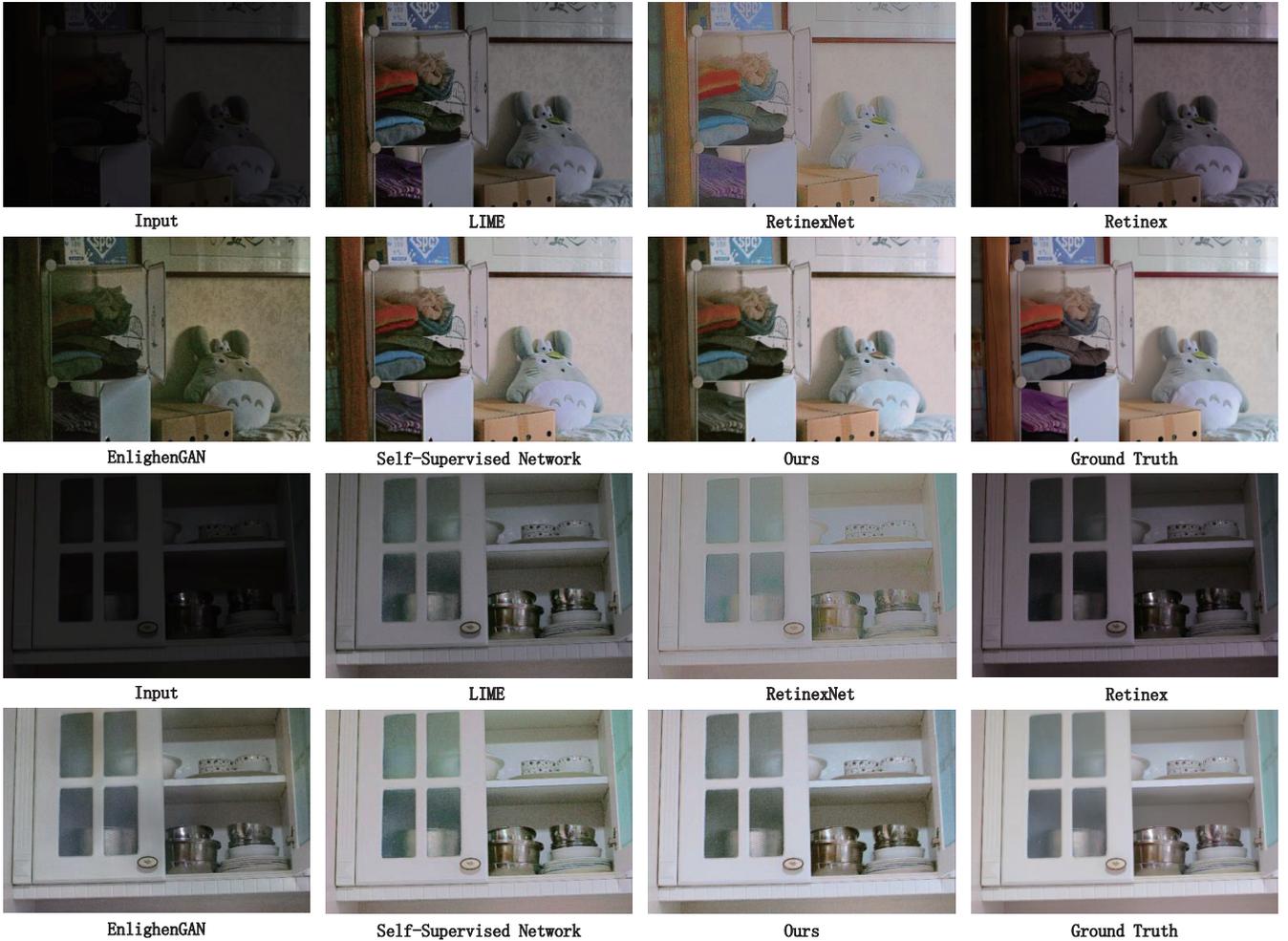}
	\caption{Visual comparison with LLIE methods on LOL-v1 dataset.}
	\label{Fig:LOLv1}
\end{figure*}

\section{Experimental Results}
\subsection{Benchmark Datasets and Experimental Details}
To evaluate the performance and efficiency of the proposed ASW-Net, we evaluated our method on some publicly available low-light datasets, including LOL (v1\cite{wei2018deep} and v2 \cite{yang2020fidelity}).
LOL-v1 dataset consists of 485 pairs of normal/low-light light training data and 15 pairs of test data.
Since our method is based on unsupervised learning, we only use the 485 low-light images for training and do not use the normal-light training images provided in the dataset.
LOL-v2 consists of two datasets, LOL-v2-real and LOLv-2-synthetic.
We conducted experiments on LOL-v2-real dataset, which includes 689 pairs of normal/low-light light training images and 100 pairs of test images.
Similarly, we only used the low-light images in the training of our network.

We quantitatively and visually compare our proposed method with several low-light image enhancement methods, including HE\cite{pizer1990contrast}, LIME\cite{guo2016lime}, SID\cite{chen2018learning}, Retinex\cite{liu2021retinex}, RetinexNet\cite{wei2018deep}, KinD\cite{zhang2019kindling}, Zero-DCE\cite{guo2020zero}, EnlightenGAN\cite{jiang2021enlightengan}, SGZ\cite{zheng2022semantic}, Self-supervised Network\cite{zhang2021self}.
We utilize the publicly available source code and recommended parameters for each of the compared methods, and fine-tune all models on the training dataset to enable comparison.
Peak Signal-to-Noise Ratio (PSNR) \cite{kingma2014adam} and Structural Similarity (SSIM) \cite{wang2004image} are adopted for quantitative comparison as evaluation metrics.

The parameters of all baseline models mentioned above are initialized as in the corresponding paper.
The value of $\lambda_1,\lambda_2,\lambda_4$ in loss function Eq.\eqref{Eq:Objective} are set to $0.1,0.1,0.01$ according to our ablation study in \ref{sec:ablation}, while $\lambda_3$ is set to $10$ which is same as in \cite{zhang2020self}.
For model training of our proposed network, we set the batch size to 4 and use the Adam optimizer with a learning rate of $10^{-3}$.
All experiments on benchmark datasets are implemented with PyTorch, on a 64 core Intel Xeon Gold 6226R CPU @2.90GHz, 256 GB memory and a Nvidia Duadro RTX 8000 GPU.

\subsection{Comparison with Baseline Methods}

\begin{table}
	\begin{center}
		\caption{Quantitative comparison on LOL-v1 dataset in terms of PSNR and SSIM. T:Traditional method; SL: Supervised Learning; UL: Unsupervised Learing.}
		\label{tab1}
		\begin{tabular}{cccc}
			\toprule
			Category & Method & PSNR & SSIM\\
			\midrule
			\multirow{3}*{T} & HE \cite{pizer1990contrast} & $14.95$ & $0.409$ \\
			& LIME \cite{guo2016lime} & $16.76$ & $0.560$ \\
			& LR3M \cite{ren2020lr3m} & $10.22$ & $0.434$ \\
			\hline
			\multirow{4}*{SL} & SID \cite{chen2018learning} & $14.35$ & $0.436$ \\
			& RetinexNet \cite{wei2018deep} & $16.77$ & $0.462$ \\
			& Retinex \cite{liu2021retinex} & $18.23$ & $0.720$ \\
			& KinD \cite{zhang2019kindling} & $17.64$ & $0.762$ \\
			\hline
			\multirow{5}*{UL} & Zero-DCE \cite{guo2020zero} & $14.86$ & $0.562$ \\
			& SGZ \cite{zheng2022semantic} & $14.89$ & $0.675$ \\
			& EnlightenGAN \cite{jiang2021enlightengan} & $17.48$ & $0.652$ \\
			& Self-Supervised Network \cite{zhang2021self} & $19.13$ & $0.651$ \\
			& Ours & $19.09$ & $0.710$ \\
			\bottomrule
		\end{tabular}
	\end{center}
\end{table}

In this section, We mainly evaluate our proposed method on the low-light image datasets LOL-v1 and LOL-v2.
On the LOL-v1 dataset, we compare our methods with a variety of both traditional and modern approaches.
These include traditional methods such as HE\cite{pizer1990contrast}, LIME\cite{guo2016lime} and LR3M\cite{ren2020lr3m}, supervised methods like SID\cite{chen2018learning}, RetinexNet\cite{wei2018deep}, Retinex\cite{liu2021retinex}, KinD\cite{zhang2019kindling}, as well as Zero-DCE\cite{guo2020zero}, EnlightenGAN\cite{jiang2021enlightengan}, SGZ \cite{zheng2022semantic}, Self-Supervised Network\cite{zhang2021self}.
Quantitative results on LOL-v1 dataset are shown in Table \ref{tab1}.
From Table \ref{tab1}, we can see that our method obtains the excellent PSNR and SSIM scores in all unsupervised methods, and is even better than part of supervised methods.

To further analyze the enhancement performance of our proposed method, we provide a visual comparison of performance between ASW-Net and baseline models on the LOL-v1 dataset in Fig. \ref{Fig:LOLv1}.
From Fig. \ref{Fig:LOLv1} we can see that images restored by LIME, Retinex or EnlighenGAN are of obviously insufficient brightness, while those enhanced by RetinexNet have lower contrast and unnatural colors.
Images enhanced by Self-Supervised Network are slightly blurry.
Compared to these baseline models, our proposed method resolve the issues of color distortion and brightness imbalance.
We can observe that the enhanced image via our method obtains more pleasing quality than the baseline methods from Fig. \ref{Fig:LOLv1}.

On the LOL-v2 dataset, we compare our methods with traditional method LIME\cite{guo2016lime}, supervised methods including SID\cite{chen2018learning}, Retinex\cite{liu2021retinex}, and unsupervised methods such as Zero-DCE\cite{guo2020zero}, EnlightenGAN\cite{jiang2021enlightengan}.
Table \ref{tab2} summarizes quantitative results on LOL-v1 dataset.
We can see from Table \ref{tab2} that our method obtains the excellent PSNR and SSIM scores compared with all unsupervised learning methods, and is even better than part of supervised learning methods.

\begin{table}
	\begin{center}
		\caption{Quantitative comparison on LOL-v2 test set in terms of PSNR and SSIM.T:Traditional method; SL: Supervised Learning; UL: Unsupervised Learing.}
		\label{tab2}
		\begin{tabular}{cccc}
			\toprule
			Category & Method & PSNR & SSIM\\
			\midrule
			T & LIME \cite{guo2016lime} & $15.24$ & $0.470$ \\ \hline
			\multirow{2}*{SL} & SID \cite{chen2018learning} & $13.24$ & $0.442$ \\
			& Retinex \cite{liu2021retinex} & $18.37$ & $0.723$ \\
			\hline
			\multirow{3}*{UL} & Zero-DCE \cite{guo2020zero} & $20.54$ & $0.781$ \\
			& EnlightenGAN \cite{jiang2021enlightengan} & $18.23$ & $0.612$ \\
			& Ours & $18.39$ & $0.639$ \\
			\bottomrule
		\end{tabular}
	\end{center}
\end{table}

\subsection{Ablation Study} \label{sec:ablation}
\begin{figure}[!ht]
	\centering
	\subfloat[Input]
	{\includegraphics[width=0.24\textwidth]{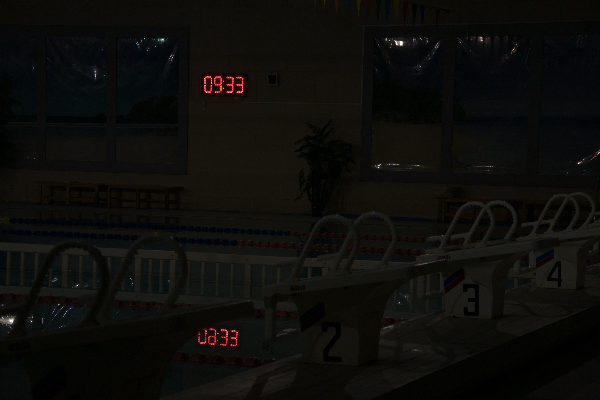} }
	\subfloat[Ground Truth]
	{\includegraphics[width=0.24\textwidth]{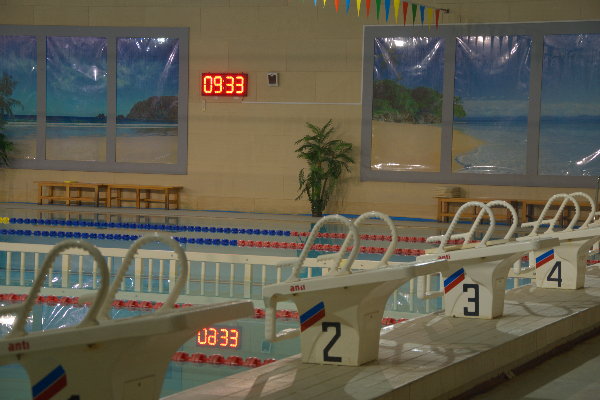}} \\
	\subfloat[SwinIR]
	{\includegraphics[width=0.24\textwidth]{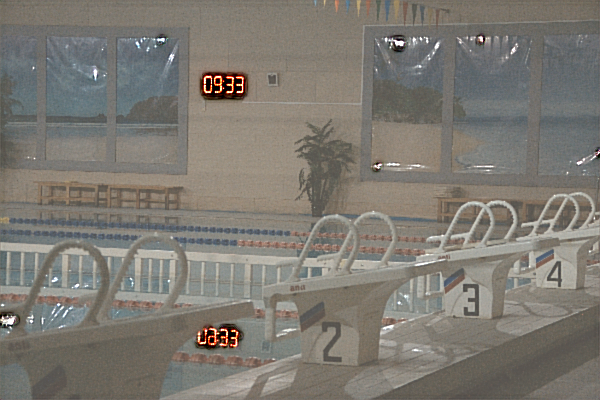}}
	\subfloat[Ours]
	{\includegraphics[width=0.24\textwidth]{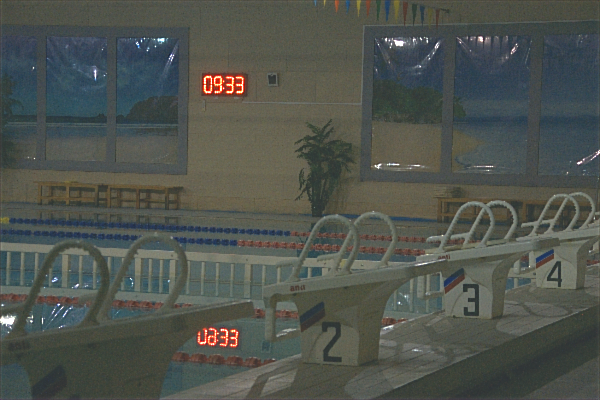}}
	\caption{Visual ablation study of the proposed method.}
	\label{Fig_ablation}
\end{figure}

In this section, we conduct ablation study to demonstrate the effectiveness of our SNR guided component and loss function.
First, we conduct experiments to evaluate the effectiveness of SNR guided component.
The setting can be find in Table \ref{tab3}.
We compare the results of our method with and without SNR guided component, i.e., SwinIR \cite{liang2021swinir}.
As show in Fig. \ref{Fig_ablation}, the SNR guided component can suppress the unpleasing artifacts, including over-exposure, noise corruption and brightness imbalance, and in Table \ref{tab3} our method achieve higher performance with respect to the metrics of PSNR and SSIM, which confirm the effectiveness of SNR guided component.

\begin{table}[!ht]
	\begin{center}
		\caption{Quantitative results of the ablation study. SNR denotes our SNR guided module.}
		\label{tab3}
		\begin{tabular}{cccc}
			\toprule
			Dataset & SNR & PSNR & SSIM\\
			\midrule
			\multirow{2}*{LOLv1} & \scriptsize{\Checkmark}  & $19.09$ & $0.710$ \\
			& \scriptsize{\XSolid}   & $18.84$ & $0.664$ \\
			\midrule
			\multirow{2}*{LOLv2} & \scriptsize{\Checkmark} & $18.39$ & $0.639$ \\
			&  \scriptsize{\XSolid} & $18.07$ & $0.612$ \\
			\bottomrule
		\end{tabular}
	\end{center}
\end{table}

\begin{figure*}[!ht]
	\centering
	\includegraphics[width=0.9\linewidth]{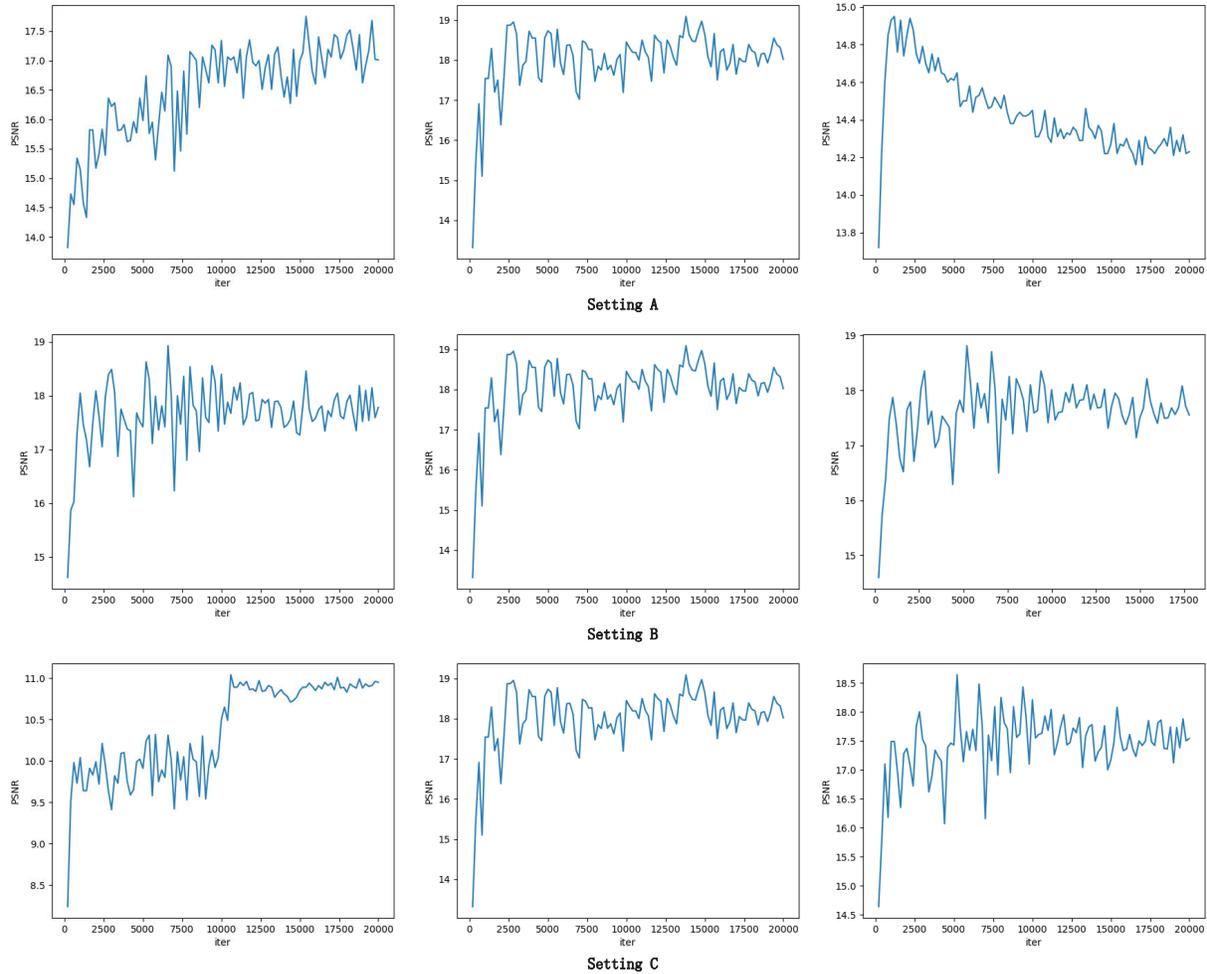}
	\caption{Evaluation indexes on the LOL-v1 testing data by different $\lambda$. Setting A: $\lambda_1=1/0.1/0.01$. Setting B: $\lambda_2=1/0.1/0.01$. Setting C: $\lambda_4=0.1/0.01/0$.}
	\label{Fig_loss}
\end{figure*}

Subsequently, in order to find the optimal setting of parameter $\lambda_1$,$\lambda_2$ and $\lambda_4$ in loss function, we set up a series of experiments for analysis and discrimination.
We analyze the advantages and disadvantages of different settings in loss function according to the enhanced image and the value of evaluation index.
The formulas of these loss functions are as follows:
\begin{equation}
	\begin{split}
		& \ell_R=\lambda_1 \left \| \max_{c\in r,g,b}  R^{c}- F \left( \max_{c\in r,g,b}  S^{c} \right) \right \|_1 +\lambda_4 \left \| \nabla R \right \| _1, \\
		& \ell_I=\lambda_2 \left \| \nabla I\circ \exp\left(-\lambda_3 \nabla R \right) \right \|_1.
	\end{split}	
\end{equation}

\begin{figure*}[!ht]
	\centering
	\includegraphics[width=1.0\linewidth]{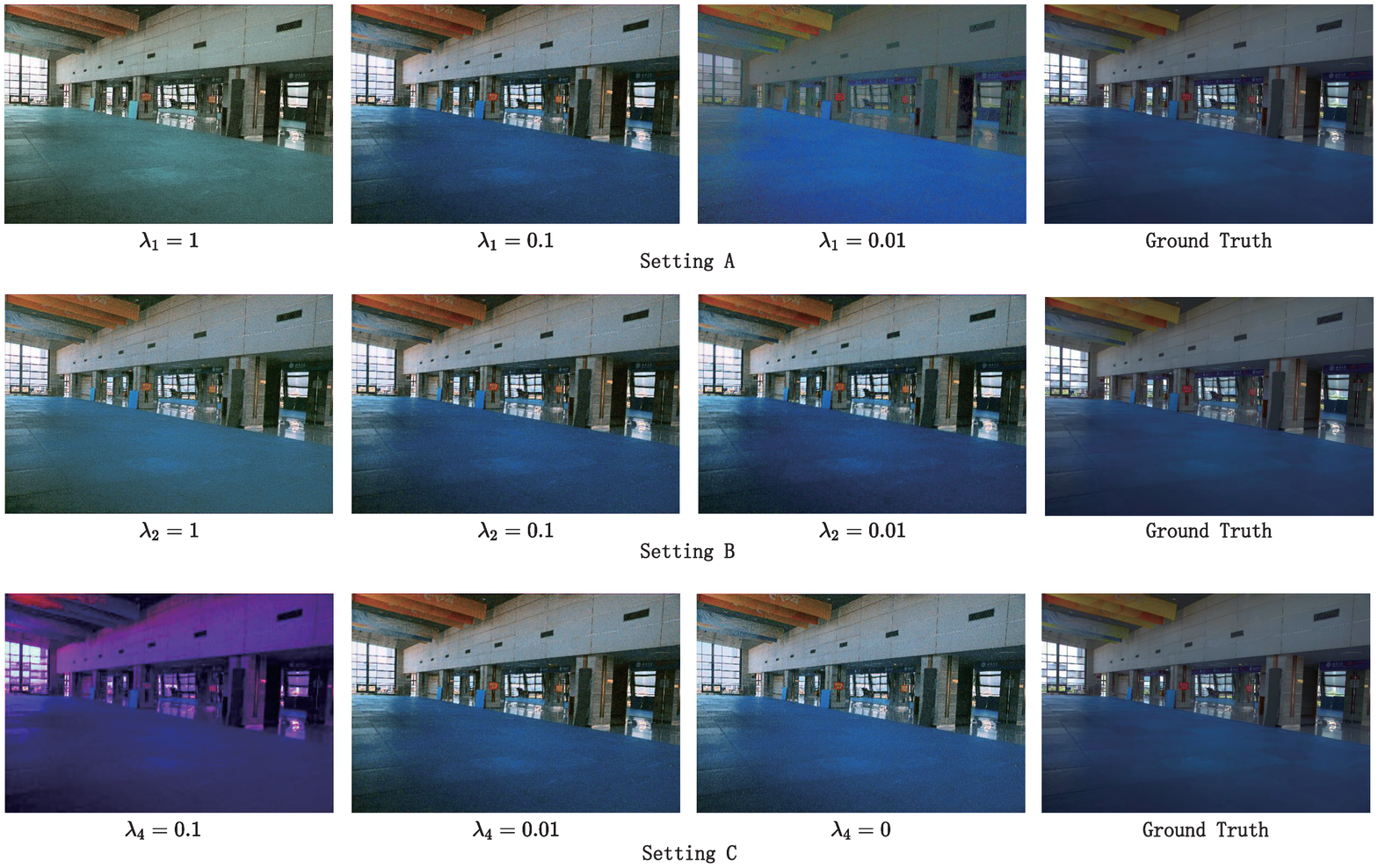}
	\caption{Evaluation indexes on the LOL-v1 testing data by different $\lambda$. Setting A: $\lambda_1=1/0.1/0.01$. Setting B: $\lambda_2=1/0.1/0.01$. Setting C: $\lambda_4=0.1/0.01/0$.}
	\label{Fig_lambda}
\end{figure*}

We set up three sets of parameters in the experiment with by varying the value of $\lambda$ while fixing others:
Setting A: $\lambda_1=(1/0.1/0.01)$, Setting B: $\lambda_2=(1/0.1/0.01)$ and Setting C: $\lambda_4=(0.1/0.01/1)$.
As show in Fig. \ref{Fig_loss}, we can observe that prior of reflectance can suppress noise and preserve the consistency of color, and the prior of illumination can be local consistent and structure aware.
We plot the evaluations of different parameter setting in Fig. \ref{Fig_loss}.
From Fig. \ref{Fig_loss} and Fig. \ref{Fig_lambda}, the parameters of $\lambda_1$,$\lambda_2$ and $\lambda_4$ in loss function are chosen as $\lambda_1=0.1,\lambda_2=0.1, \lambda_4=0.01$.

\section{Conclusion}
In this paper, we proposed a dual branch network with SNR-aware Swin Transformer for low-light image enhancement, where an SNR-aware transformer module is designed in a spatial-varying manner.
To eliminate the dependence of data collection and to generalize the network ASW-Net to be extended to enhance images collected from different scenes, different lighting conditions and different devices, we build an unsupervised framework based on Retinex model, possessing more feasible and efficient training of network.
Experimental results demonstrate that the proposed model is competitive with the baseline model.

\bibliographystyle{plain}
\bibliography{ASW-Manuscript}

\end{document}